\documentclass{article}

\PassOptionsToPackage{round}{natbib}

\usepackage[preprint]{neurips_2025}

\usepackage{tcolorbox}
\usepackage{tikz}
\usetikzlibrary{shadings}
\usetikzlibrary{arrows.meta, positioning}
\usepackage{soul}
\usepackage[utf8]{inputenc} 
\usepackage[T1]{fontenc}    
\usepackage{hyperref}       
\usepackage{url}            
\usepackage{booktabs}       
\usepackage{amsfonts}       
\usepackage{nicefrac}       
\usepackage{microtype}      
\usepackage{xcolor}         
\usepackage{amsmath}
\usepackage{graphicx}
\usepackage{cleveref}
\usepackage{algorithm}         
\usepackage{algpseudocode}     
\usepackage{amsmath}           

\usepackage{subcaption}

\usepackage{enumitem}
\setlist[itemize]{leftmargin=2em}  

\usepackage{xcolor}
\definecolor{color1}{rgb}{0.298, 0.447, 0.690}  
\definecolor{color2}{rgb}{0.866, 0.518, 0.322}  
\definecolor{color3}{rgb}{0.333, 0.659, 0.408}  
\definecolor{color4}{rgb}{0.769, 0.306, 0.322}  
\definecolor{color5}{rgb}{0.506, 0.447, 0.702}  
\definecolor{color6}{rgb}{0.576, 0.471, 0.376}  
\definecolor{color7}{rgb}{0.855, 0.545, 0.765}  
\definecolor{color8}{rgb}{0.549, 0.549, 0.549}  
\definecolor{color9}{rgb}{0.8, 0.725, 0.455}    
\definecolor{color10}{rgb}{0.392, 0.710, 0.804} 

\pgfdeclareverticalshading{rainbowshading}{100bp}{
  color(0bp)=(color5!70!white);
  color(30bp)=(color5!70!white);
  color(35bp)=(color4!70!white);
  color(50bp)=(color3!70!white);
  color(65bp)=(color2!70!white);
  color(70bp)=(color1!70!white);
  color(100bp)=(color1!70!white)
}

\definecolor{RoyalBlue}{rgb}{0.0, 0.2, 0.6}
\hypersetup{
  colorlinks=true,
  linkcolor=RoyalBlue,     
  citecolor=RoyalBlue,     
  urlcolor=RoyalBlue       
}

\title{Group Think: Multiple Concurrent Reasoning Agents Collaborating at Token Level Granularity}

\author{%
  \begin{tabular}[t]{c}
    Chan-Jan Hsu  ~~ Davide Buffelli ~~ Jamie McGowan \\ \\
    Feng-Ting Liao ~~ Yi-Chang Chen ~~ Sattar Vakili ~~ Da-shan Shiu 
  \end{tabular}
  \\ \\
    MediaTek Research\thanks{Correspondence to Chan-Jan Hsu <\texttt{chan.hsu@mtkresearch.com}>.}
}

\begin{document}

\maketitle

\begin{abstract}
Recent advances in large language models (LLMs) have demonstrated the power of reasoning through self-generated chains of thought. Multiple reasoning agents can collaborate to raise joint reasoning quality above individual outcomes. However, such agents typically interact in a turn-based manner, trading increased latency for improved quality.
In this paper, we propose \textit{Group Think}---a single LLM that acts as multiple concurrent reasoning agents, or thinkers. With shared visibility into each other's partial generation progress, Group Think introduces a new concurrent-reasoning paradigm in which multiple reasoning trajectories adapt dynamically to one another at the \textit{token level}. For example, a reasoning thread may shift its generation mid-sentence upon detecting that another thread is better positioned to continue.
This fine-grained, token-level collaboration enables Group Think to reduce redundant reasoning and 
improve quality while achieving significantly lower latency. Moreover, its concurrent nature allows for efficient utilization of idle computational resources, making it especially suitable for \textit{edge inference}, where very small batch size often underutilizes local~GPUs.
We give a simple and generalizable modification that enables any existing LLM to perform Group Think on a local GPU. We also present an evaluation strategy to benchmark reasoning latency and empirically demonstrate latency improvements using open-source LLMs that were not explicitly trained for Group Think. We hope this work paves the way for future LLMs to exhibit more sophisticated and more efficient collaborative behavior for higher quality generation.
\end{abstract}

\section{Introduction}
\label{sec:intro}

Large language models (LLMs) have achieved strong results across a wide range of complex natural language tasks~\citep{minaee2025largelanguagemodelssurvey}. 
Recent developments have increasingly focused on enhancing the \emph{reasoning} capabilities of LLMs by leveraging additional \emph{test-time compute}~\citep{ji2025test}. These approaches allocate more computation during inference to improve model performance on challenging tasks.
A key direction in this space involves requesting models to generate a sequence of intermediate reasoning steps---referred to as chain-of-thought (CoT)~\citep{CoT}---prior to producing a final answer. More recently, models have been trained to self-generate CoT reasoning without explicit prompting ~\citep[e.g.,][]{openAIo1,deepseekai2025deepseekr1,geminiThinking,zhao2024marcoo1openreasoningmodels,qwenQWQ}.

Another active area of research focuses on \emph{agentic} systems powered by LLMs~\citep{li2024survey}. These systems use LLMs to drive agents that can autonomously make decisions and take actions based on their environment. The community is increasingly exploring settings where multiple reasoning agents collaborate to achieve a shared objective~\citep{tran2025multi}. Recent work has demonstrated that distributing subtasks among agents—and enabling efficient, and effective, communication and collaboration—is key to unlocking their full potential~\citep{chen2024agentverse,feng2025one}.

Many works have explored different mechanisms of collaboration between agents~\citep{piatti2024cooperate, tran2025multi}. Most current methods are built upon \emph{turn-based} communication; that is, agents exchange messages back and forth in a \emph{chat chain}~\citep{qian-etal-2024-chatdev,chen2024reconcile}. This turn-by-turn mechanism can introduce significant latency due to the sequential nature of the protocol. However, in a turn-based setting, it is non-trivial for agents to work in a concurrent manner efficiently. An agent that speculatively generates its outputs and actions can often discover that, at the end of turn, its work is duplicative of, inconsistent with, or even contradictory to other agents'. 

Inspired by the potential of multi-agent systems, we introduce a new generation paradigm in which one LLM produces, in parallel, multiple \emph{communicating} reasoning trajectories. We call this method \emph{Group Think}, as it resembles the process of multiple individuals working together toward the solution of a task. Unlike typical multi-agent systems, mechanistically, Group Think can operate at token-level granularity and can be fully integrated into the model's inference procedure without any further framework.

By enabling multiple concurrent reasoning threads that adapt token-by-token to the content generated by others, Group Think raises generation quality and, at the same time, reduces generation redundancy and latency. We test Group Think on open source LLMs. Our results show that current models, while not explicitly trained for the Group Think paradigm, can already take advantage of this generation mechanism, although not to its full potential. 

In addition to higher quality, a distinct advantage of Group Think is its particularly compelling efficiency in the context of \textit{local} LLM inference. Local deployments, especially on personal devices, typically operate with a batch size of one. Such a small batch size unfortunately only makes use of a minute portion of the capability of modern processors, as an entire system is bottlenecked by transferring model weights from the memory to the processor.  Group Think can put this idle capacity to use by running multiple concurrent reasoning threads. This can result in a dramatic reduction in the latency overhead of reasoning, making the deployment of small language model on local devices significantly more feasible just as the quality of small reasoning LLM starts to approach practical use. 

In our experimental we carefully quantify the latency improvements provided by Group Think. To gain clear insights into how Group Think trades concurrent throughput for latency, we measure generation latency in tokens across a few selected prototypical problem classes. We compare the latency of an LLM with and without Group Think. Furthermore, we evaluate accuracy-latency trade-offs when generating multiple independent candidate solutions concurrently~\citep{brown2024large,wang2023selfconsistency}. Since independent reasoning can be seen as a special case of Group Think, it serves as a natural floor for the performance-latency trade-off.

In summary, our contributions are:
\begin{itemize}
\item[$\diamond$] We introduce {Group Think}, a new generation paradigm in which an LLM produces multiple interdependent, parallel reasoning trajectories—resembling a group of individuals collaboratively solving a problem.
\vspace{-1mm}
\item[$\diamond$] We present two efficient implementation methods for Group Think, highlighting the advantages it brings to edge inference 
\vspace{-1mm}
\item[$\diamond$] We evaluate Group Think across several representative problems. Our results show that Group Think enables LLM to achieve improved reasoning accuracy while reducing latency. These results demonstrate that existing LLMs already exhibit emergent support for this paradigm.

\end{itemize}

We note that current LLMs are not explicitly trained to perform Group Think. As such, the results presented in this paper should be viewed as an auspicious starting point. We believe that by training models on dedicated collaborative reasoning data, the benefits of Group Think could be significantly amplified. We invite the research community to join us in developing such datasets.

\section{Related Work}
\label{sec:rel_work}

\paragraph{Sequential single-agent reasoning methods.} In single agent settings, increasing the amount of computation performed at inference time has been incredibly successful at delivering state-of-the-art performance on complex tasks 
 \citep{ji2025test,zhang2025and}. This extra computation can be leveraged in a variety of ways in language modelling, such as encouraging LLMs to critique and refine responses \citep{kumar2025training}, or pause to deliberate during generation \citep{goyal2024think}. One of the most popular approaches, which is tightly related to our work, relies on asking the model to provide a solution step by step in what is referred to as Chain-of-Thought \citep[CoT,][]{CoT}, where the LLM is permitted to generate a sequence of intermediate steps/thoughts before generating the final answer. This approach showed significant advantage over more naive prompting techniques.

\paragraph{Multiple independent single-agent reasoning methods.} The beneficial effect of a single CoT can be amplified by allocating further inference-time compute to multiple sampling of CoTs. Naive multiple sampling involves sampling multiple solutions independently, then selecting the best one either using a reward model \citep{brown2024large}, or a majority voting approach (self-consistency) \citep{wang2023selfconsistency}. Being quite simple to implement, this technique has been shown to lead to significant performance gains when scaled up \citep{singhi2025solve,zhao2025sample}. 

To reduce the overall overhead of generating reasoning chains, an extension to the aforementioned multiple independent CoTs concept was proposed by \citet{10.5555/3666122.3666639} and \citet{long2023large} termed Tree-of-Thoughts, which enhances the thought generation by introducing a tree-like structure. This is achieved by allowing new trajectories/branches to spawn from profitable reasoning traces. Subsequent work by \citet{besta2024graph} introduced Graph-of-Thoughts (GoT) to allow further flexibility of inheriting generated thoughts into newly spawned nodes, thus enabling new nodes to define a new chain or to augment existing ones. 

\paragraph{Sequential multi-agent methods.} With the growing adoption of LLMs in practical applications, significant attention has been devoted to developing agentic systems—particularly those composed of multiple LLM-based agents, each executing its own chain-of-thought (CoT) reasoning. The literature on multi-agent LLM systems is extensive, encompassing a range of communication paradigms, including cooperative frameworks \citep{hongmetagpt, chen2024reconcile}, competitive setups \citep{liang2024encouraging}, role-based  \citep{chenagentverse} and rule-based coordination mechanisms \citep{xu2023towards}, and decentralized models \citep{zhang2023cumulative}. For a comprehensive overview of this fast-evolving area, we refer to recent survey articles \citep{tran2025multi, ijcai2024p890, li2024survey}.

\paragraph{Concurrent multi-agent generation.} Several works have proposed methods to make the generation procedure of LLMs or agents concurrent. \citet{wang2024mixture} introduce the use of \emph{mixture of agents}, i.e., concurrent agents that communicate at regular intervals. \citet{zhuge2024gptswarm} introduce a framework for orchestrating agents in a chain-like sequential process. Recent works \citep{ningskeleton,kim2024llm,pan2025learningadaptiveparallelreasoning,jin2025learning} have further extended the research direction of concurrent generation by introducing \emph{dynamic} methods in which the LLM is trained or prompted to decide when to perform certain tasks concurrently during generation. These works operate with independent concurrent traces, while we introduce interdependencies between concurrent traces through token-level adaptation.

\paragraph{Parallel generation in LLMs with shared latents.} Concurrent work \citep{rodionov2025hogwild} has proposed an inference method called ``Hogwild!'' with parallel reasoning traces generated concurrently (similarly to Group Think) and with shared Key-Value (KV) cache. This work is  complementary to ours as it focuses on different strategies to allow communication across reasoning traces and improve caching, while our work discusses the benefits for local execution of LLMs (in small batch settings).

\section{Group Think}
\label{sec:method}

\begin{figure}[t]
    \centering
    \begin{minipage}[t]{0.4\textwidth}
        \vspace{0pt} 
        \includegraphics[width=\textwidth]{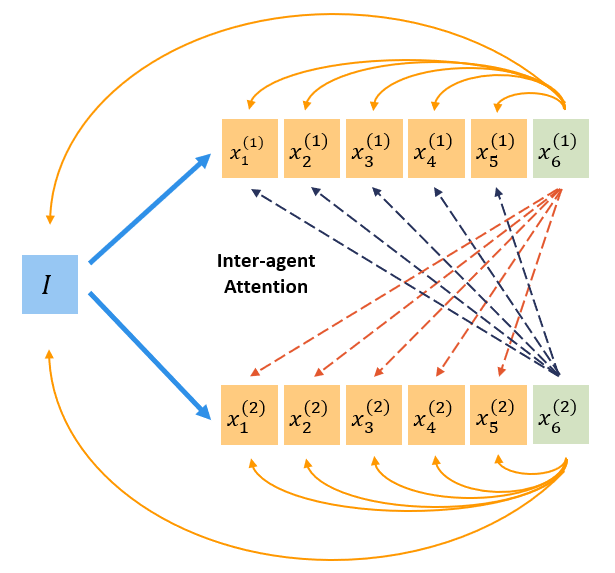}
    \end{minipage}
    \hfill
    \begin{minipage}[t]{0.5\textwidth}
        \vspace{20pt} 
        \captionof{figure}{\textit{Illustration of Group Think.} Distinct threads of reasoning thoughts (indicated with superscripts ${(1)}, {(2)}$) are parallelized with an inter-agent attention mechanism acting at the token level. Each token $x_k^{(\cdot)}$ in a thread is able to attend to all previous tokens from all other threads at every generation time step, thereby inducing fine-grained collaborative behaviour. Tokens in green are generated in parallel. Normal intra-agent self-attention are depicted by solid orange arrows, while inter-agent cross attention by dashed lines.}
        \label{fig:gt-main}
    \end{minipage}
\end{figure}

We formally define Group Think as a reasoning paradigm in which multiple synchronized chains of thought (CoT) are generated by parallelized agents (or \emph{threads}), with each agent’s reasoning process dynamically adapting to the evolving thoughts of others at the token level (see Figure \ref{fig:gt-main}). Throughout this section, we draw an analogy between LLM threads and human collaborators jointly solving a problem, and use the terms \emph{thread}, \emph{thinker}, and \emph{agent} interchangeably.

\subsection{Preliminaries}

\paragraph{Single-Agent Chain-of-Thought Reasoning.}
In standard single-agent CoT reasoning---used by models such as DeepSeek R1 \citep{deepseekai2025deepseekr1} and GPT-o1 \citep{openAIo1}---the model is prompted with an input $I$ and generates an explicit intermediate CoT $X = \{x_{1}, \dots, x_{K}\}$ of length $K$, before producing the final answer $Y$. This two-step process can be formalized as
\[
X = \text{Think}(I), \quad Y = \text{Answer}(I, X).
\]
Here, $\text{Think}(I) \sim p_{\theta}(\cdot \mid I)$ with
$
p_{\theta}(X \mid I) = \prod_{k=1}^{K} p_{\theta}(x_{k} \mid x_{1:k-1}, I),
$
and $\text{Answer}(I, X) \sim p_{\theta}(\cdot \mid I, X)$ represent sampling from the same underlying model $p_{\theta}$, invoked with different prompts or decoding strategies to elicit either an intermediate reasoning trace $X$ or a final answer $Y$ conditioned on $X$.
This approach has been shown to significantly improve reasoning accuracy \citep{CoT} over non-reasoning baselines, albeit at the cost of increased inference compute and latency due to the additional generation of $X$.

\paragraph{Multi-Agent Turn-Based Chain-of-Thought Reasoning.}
To further enhance performance, multi-agent systems can reason collaboratively by sequentially exchanging complete CoT in a turn-based manner \citep{qian-etal-2024-chatdev}. In this setting, each agent $n$ generates a full CoT $X^{(n)}$ conditioned on the prompt $I$ and the chains of thought produced by previous agents:
\[
X^{(n)} = \text{Think}(I, X^{(1)}, X^{(2)}, \ldots, X^{(n-1)}).
\]
After $n$ turns, the final answer is generated based on the prompt and all collected CoT:
\[
Y = \text{Answer}(I, X^{(1)}, X^{(2)}, \ldots, X^{(n)}).
\]
While this collaborative approach can yield higher-quality reasoning, it incurs additional inference cost and latency due to the use of multiple turns.

\subsection{Token-Level, Mutually Adaptive Multi-Agent Reasoning}
\label{sec:token-level-multi-agent-reasoning}

Group Think advances prior approaches by enabling multiple agents to reason concurrently, with each agent’s chain of thought adapting at every token step to the evolving outputs of others. Let $X^{(n)}_{k} = \{x^{(n)}_{1},\dots,x^{(n)}_{k}\}$ denote the first $k$ tokens generated by agent $n$.
In Group Think, each agent’s next token prediction $x^{(n)}_{k+1}$ is conditioned on the prompt and the partially completed thoughts of all agents:
\begin{equation}
x^{(n)}_{k+1} = \text{Think}^{(n)}_{k}\left(I, X_k^{(1)},  ..., X_k^{(N)} \right),
\label{eq:group-think-1}
\end{equation}
where $k$ is the current sequence length, $\text{Think}^{(n)}_k$ emits one thinking token for agent $n$, and $N$ is the total number of concurrent agents. 
The final answer is produced by aggregating the prompt and all chains:
\begin{equation}
Y = \text{Answer}(I, \{\, X^{(n)} \mid 1 \leq n \leq N \,\}).
\label{eq:group-think-2}
\end{equation}

The actual implementation can deviate from the definition in \eqref{eq:group-think-1} for practical reasons, for instance, to fit better with a particular AI accelerator or library. See \Cref{subsec:efficient-imp}. 

This token-level, mutually adaptive reasoning enables fine-grained collaboration, allowing agents to dynamically adjust their reasoning trajectories in response to the evolving context provided by their peers. Group Think thus offers a powerful and general framework for collaborative reasoning, with the potential to achieve high-quality results with improved efficiency.

We remark that the multi-agent reasoning chains that Group Think can produce is quite expressive, subsuming those from artificially structured approaches such as Tree of Thought~\citep{10.5555/3666122.3666639,long2023large} and Graph of Thought~\citep{besta2024graph}. 
With Group Think, such trajectories can naturally emerge where profitable.

\vspace{-1mm}
\subsection{Efficient Implementation for Group Think Inference}
\label{subsec:efficient-imp}

We present two practical implementations for Group Think. One is tailored to single-query scenarios, prevalent in on-device settings where queries are typically processed individually. The other is designed for data centers to process a batch of mixed standard and Group Think requests as a unit. 

\begin{figure}
    \centering
    \includegraphics[width=0.88\textwidth]{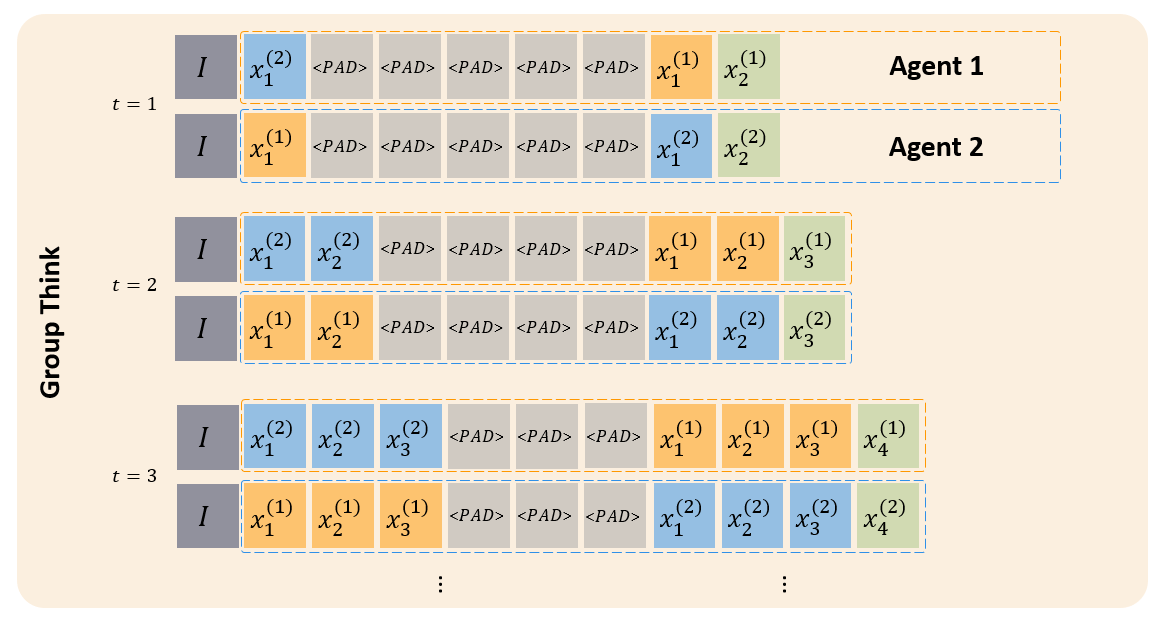}
    \caption{\textit{Implementation of Group Think for local inference scenario.} Artificial batches can be created in the low batch regime by rearranging sequences for the agents into distinct ``data points''. At each time step, sequences for different thinkers are augmented with previous predictions from other thinkers, allowing self-attention mechanisms to include cross-agent dependencies at the token level. Green tokens are generated using a standard causal mask---with some sparsity in the positional encodings corresponding to the locations of padding tokens.}
    \label{fig:gt-local-implementation}
    \vspace{-4mm}
\end{figure}
\textbf{Group Think with multiple parallel threads for the local inference scenario.} In personal or edge computing environments, inference requests are typically handled with a batch size of 1, often resulting in underutilized hardware due to memory bandwidth constraints. The Group Think inference method addresses this limitation and maximizes resource utilization: for a single query, the $N$ agents of our Group Think paradigm operate in parallel, creating an effective \emph{agent-level batch} of size $N$. As long as weight retrieval through the memory interface stays as the system bottleneck, running Group Think in this way incurs no additional latency.

This method is illustrated in \Cref{fig:gt-local-implementation} for $N=2$. Each agent is allocated a token budget $K$. After the prompt, each agent $n$ generates its next token $x^{(n)}_k$ in parallel. 
For each agent, a sequence of $K(N-1)$ is dedicated to the tokens of previous agents, and each new token $x^{(\cdot)}_k$ gets assigned to position index $K(N-1) + k$.
To allow each agent to access the KV values for tokens generated by other agents, this implementation modifies the standard causal attention mask such that agents attend to both the shared prompt and the tokens generated thus far across all agents, with some sparsity to account for the positions assigned to the tokens of other agents that are not yet filled in. 

\begin{figure}
    \centering
    \includegraphics[width=0.95\textwidth]{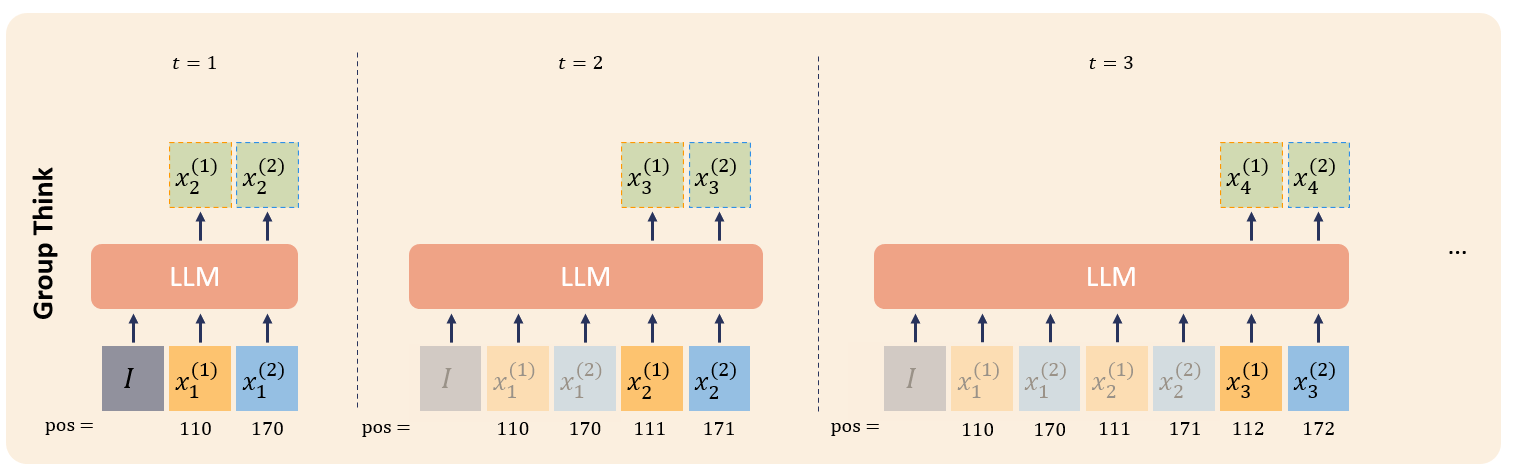}
    \caption{\textit{Implementation of Group Think for single thread data center inference scenario.} Each agent is allocated a slot of token indexes: agent 1 (orange tokens) is allocated positions 110 to 169, while agent 2 (blue tokens) has 170 to 219. Token prediction for all agents is performed by adapting the sequence layout to interleave tokens from different agents, hence leading to non-sequential positional indices. At each time step $t$, $N$ new tokens (green) are generated with a common causal mask, allowing both intra-agent and inter-agent attention. Tokens from previous timesteps (represented by semi-transparent blocks) are inserted in the KV cache. This design allows Group Think instances to be processed together with other requests in a batch within a data center environment.}
    \label{fig:gt-implementation}
    \vspace{-2mm}
\end{figure}

\textbf{Group Think with a single thread for the data center scenario.} For data center applications, it is generally required that multiple requests are aggregated into a single batch for processing\footnote{In data center environments, computational efficiency is maximized by processing multiple user requests simultaneously in batched operations.}---whether employing Group Think or not. To enable this, we show that it is possible to take advantage of Group Think with a single thread through a simple modification to the generation procedure.  

The core insight enabling Group Think on a single thread is the token-by-token interleaving of generation among agents during inference, as illustrated in \Cref{fig:gt-implementation}. 
More precisely, each agent is allocated a \emph{slot} of token indices (which determine the corresponding positional embeddings), and each generation step fills one token per agent, leading to a KV cache \citep{luohe2024keep} with interleaved tokens from each agent\footnote{In this setting, the causal ordering of the KV cache typically does not faithfully follow the positional ordering of the token sequence.}. By interleaving the tokens across agents in this way, the causal mask in the attention mechanism allows each new token to attend to all previously generated tokens (which includes tokens from all agents), thus realizing the benefits of Group Think without any architectural modification.
A key aspect of this implementation is that the token generation order is decoupled from the positional indices: tokens assigned to earlier positions may attend to later-positioned tokens (from other agents), if those have already been generated. We provide additional details in \Cref{app:data_center_implementation}.

Crucially, because the underlying attention mechanism is not modified, this method allows for the multiplexing of standard and Group Think inference requests within the same processing batch. Furthermore, this interleaving principle can be extended to the training phase. Training data can be prepared by formatting sequences to include contributions from multiple agents, each associated with its designated (and potentially non-contiguous) position indices. This allows for the inclusion of Group Think-style instances alongside standard data in training batches, thereby enabling the reuse of existing transformer training frameworks with minimal modification.

\section{Can Pretrained LLMs Elicit Group Think? An Evaluation}
\label{sec:experiments}

In this section, we present an empirical study to assess whether existing models already exhibit the ability to leverage the Group Think paradigm. We focus on three problem categories chosen to provide insight into the potential behaviors and benefits of Group Think: \emph{enumeration}, \emph{divide and conquer}, and \emph{coding}. We provide general prompt and compute details in Appendix \ref{app:general_exp_details}.

For each problem category, we evaluate the performance–latency trade-off by measuring the \emph{completion coverage} of the solution at various per-thinker generation length, measured in number of tokens per thinker. With a reasonable hardware and software implementation, we expect the real-world latency to be largely proportional to the longest generation length among multiple agents. Therefore, we adopt the per-thinker generation length to represent latency. We use off-the-shelf, instruction-tuned language models in our experiments. Since these models are not explicitly trained for Group Think, the results reported here provide a conservative lower bound on its potential benefits.

We begin by comparing Group Think to standard CoT reasoning, the baseline inference method for recent instruction-tuned models. We then evaluate it against multiple concurrent Independent Sampling (IS) ~\citep{brown2024large}. This baseline can be viewed as a special case of Group Think where reasoning threads evolve independently, without visibility into each other’s progress. This contrast allows us to isolate and quantify the benefits specifically due to thinker collaboration. 

\begin{figure}
    \begin{subfigure}[t]{0.325\textwidth}
      \includegraphics[width=\textwidth]{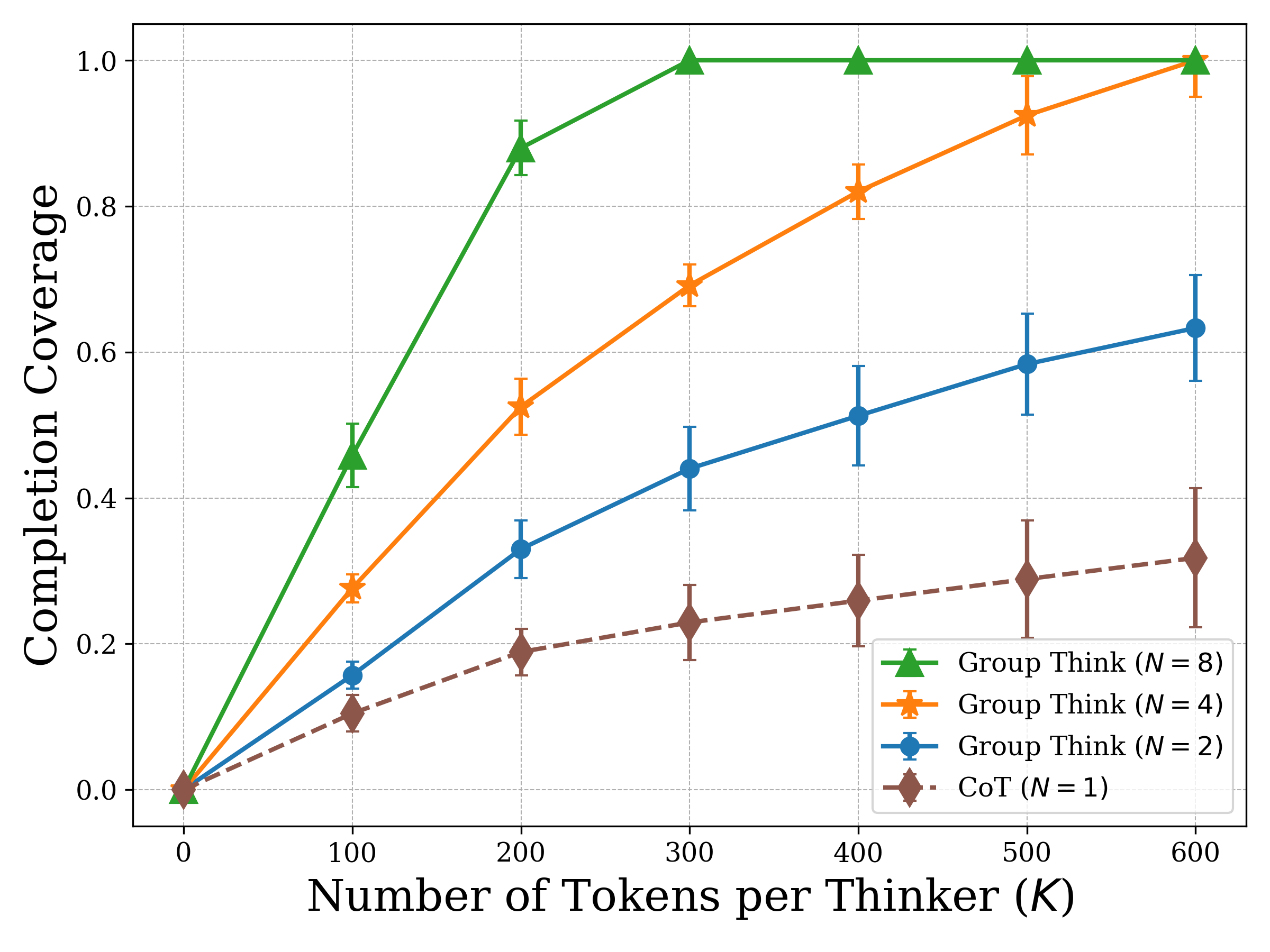}
      \caption{Enumeration}
      \label{fig:enum-results}
    \end{subfigure}
    \hfill
    \begin{subfigure}[t]{0.325\textwidth}
      \includegraphics[width=\textwidth]{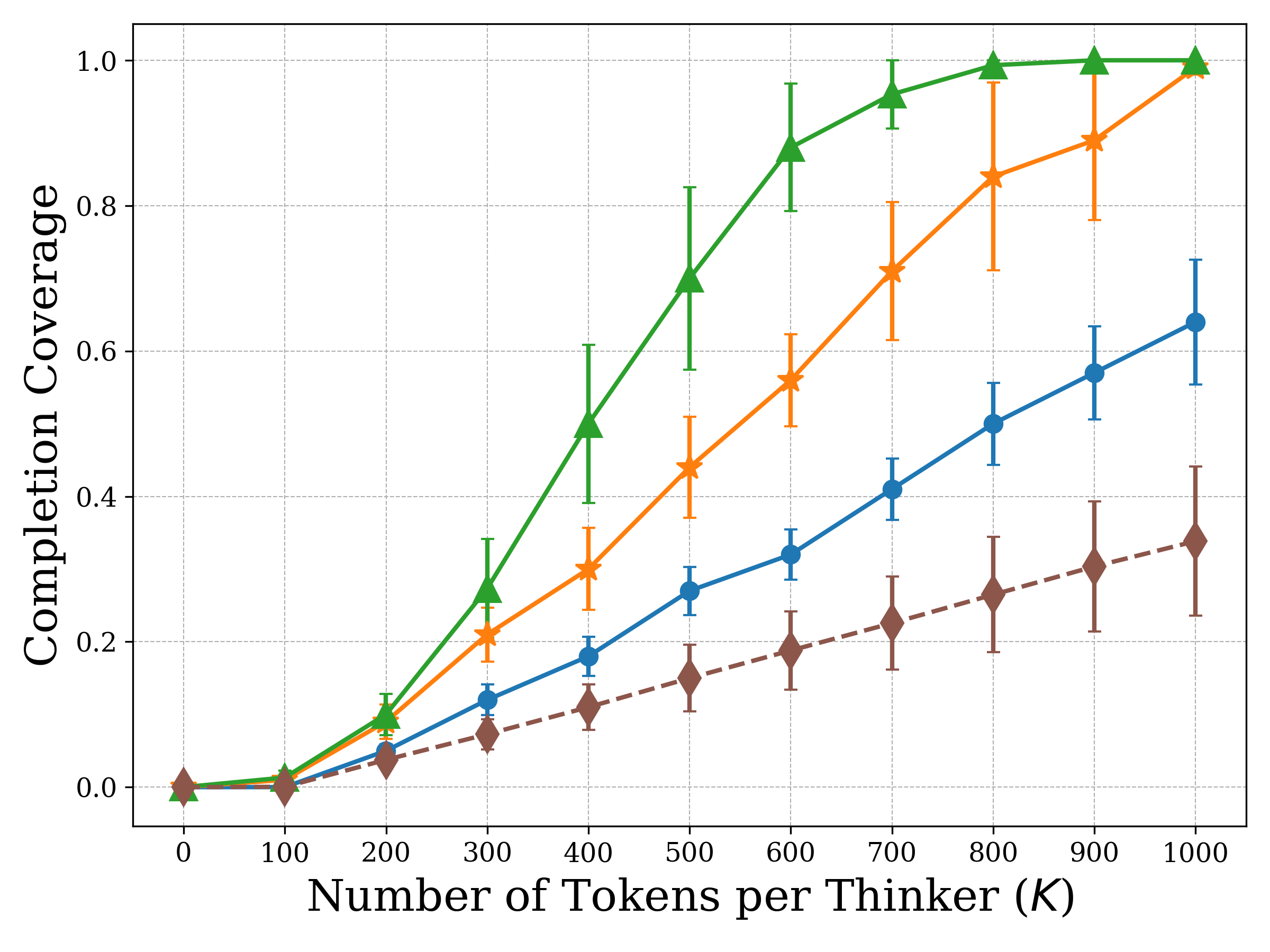}
      \caption{Divide and Conquer}
      \label{fig:div-results}
    \end{subfigure}
    \hfill
    \begin{subfigure}[t]{0.325\textwidth}
      \includegraphics[width=\textwidth]{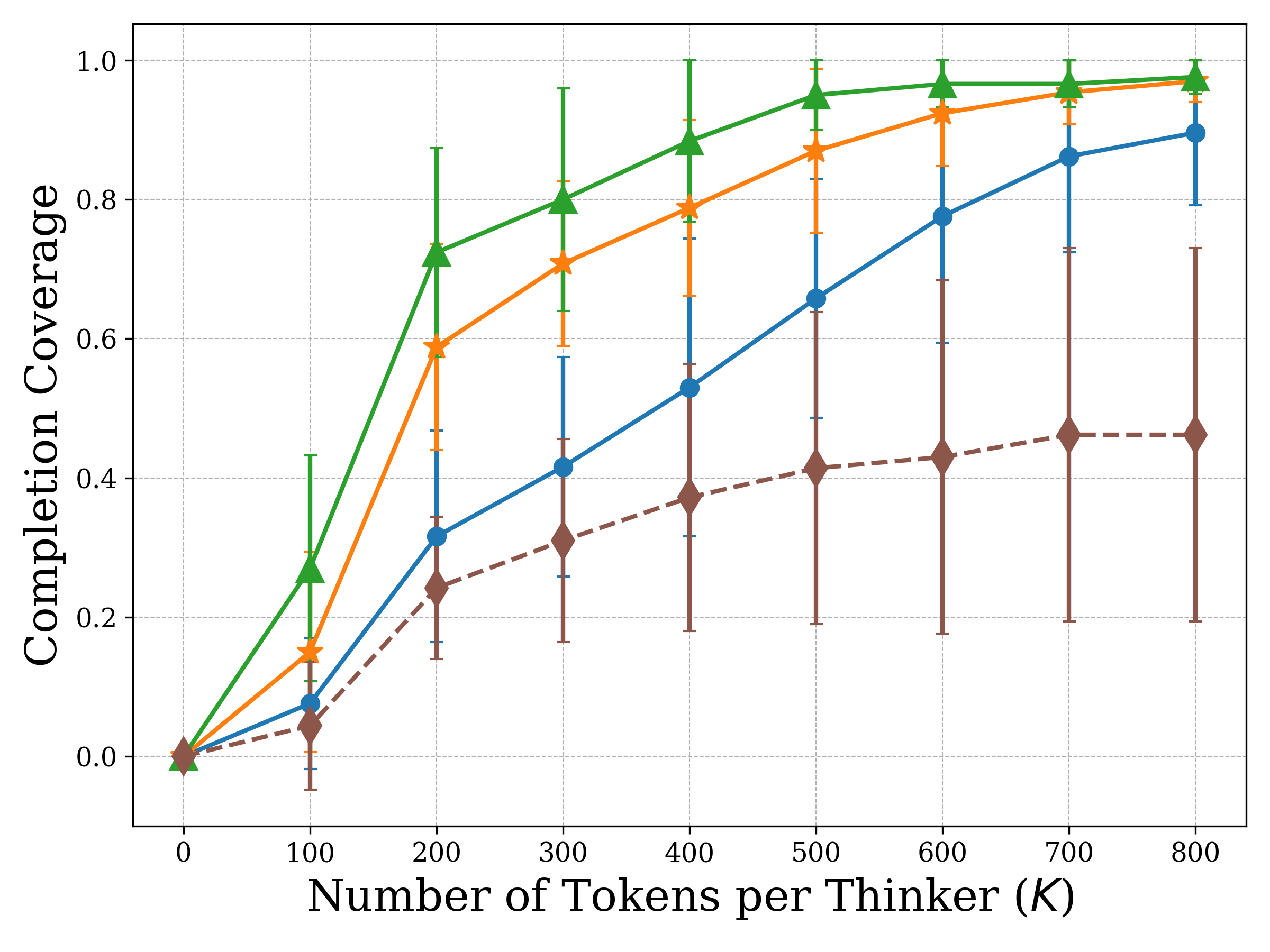}
      \caption{Programming}
      \label{fig:code-results}
    \end{subfigure}
    \caption{Completion coverage vs. latency comparing Group Think to the CoT baseline. Across all tasks, Group Think starts out with an acceleration roughly $N$ (number of thinker) times faster than CoT until the Completion Coverage becomes near saturated. More thinkers always solve the problem faster. In Programming, while CoT stays far from solving the problem, Group Think with over $4$ thinkers can often get to a solution. Error bars indicate standard deviation across multiple runs.}
\end{figure}

\subsection{Enumeration}
In an enumeration task, a generation strategy (CoT, IS, or Group Think) is asked to produce a reasoning trajectory that contains $L$ distinct items from a specific category (e.g., animals, colors, or countries). While seemingly trivial, this is a fundamental skill underlying how Group Think can address real world problems efficiently. We define the completion coverage as the degree to which the joint reasoning from the Group Think agents solves the enumeration problem. This quantity can be computed analytically by counting the number of distinct items generated, normalized by~$L$.
That is,
$
\text{Completion Coverage} = \min\left\{1,\
\frac{\#\text{distinct items generated}}{L}\right\}.
\label{coverage_score_enum}
$

We construct 10 enumeration prompts spanning diverse domains and report the performance across multiple runs; e.g., \textit{``List 100 male names''}. We use \texttt{Llama-3.1 8B Instruct}\footnote{\url{https://huggingface.co/meta-llama/Llama-3.1-8B-Instruct}}. Results are shown in Figure~\ref{fig:enum-results}. As predicted, we observed that Group Think initially outperforms CoT by a factor close to $N$. This acceleration gradually slows down as the Group approaches solving the problem. Furthermore, we note that more thinkers always solve this problem faster. Additional details and prompt examples are provided in Appendix~\ref{app:eval_prompts_enum}.

The model we used to generate reasoning (\texttt{Llama-3.1 8B Instruct}) was never trained on any group collaboration strategies. In the prompt, we only instruct the thinkers to avoid duplicating what other thinkers will generate. By examining the generated reasoning trajectories, we observe emergent behaviors indicative of Group Thinking, in which agents avoid duplication by observing and adapting to the outputs of others with overt communication. For example, the agents divide the category space into meaningful subcategories and work within those—without being explicitly prompted to do so. As an illustration, in the male names example, thinkers progressively diversify their contributions by focusing on names from different cultural, historical, or regional origins—for instance, names from English-speaking countries, ancient Greek and Roman cultures, and various Asian cultures such as Chinese and Japanese. See \Cref{app:example_enum} for details of the output example. These behaviors emerge without explicit instructions in prompt, suggesting that off-the-shelf LLMs already possess foundational capabilities for collaborative reasoning under the Group Think paradigm. 

\subsection{Divide and Conquer}

Divide-and-conquer is a widely adopted problem solving paradigm. We consider it to be particularly compatible with Group Think, as it naturally decomposes a problem into smaller subproblems that can be solved independently and then aggregated into a global solution. We note that the divide-and-conquer approach requires the enumeration capability.

To evaluate how well Group Think can solve a problem via divide-and-conquer, we consider a classic Computer Science textbook problem: computing the shortest paths between all pairs of nodes in a directed, weighted graph using the Floyd–Warshall algorithm~\citep{10.1145/367766.368168}. This algorithm provides a structured setting under which to assess how, and to what extent, the thinkers progressively fill in the update space of a distance matrix of size $|\mathcal{V}| \times |\mathcal{V}|$, where $\mathcal{V}$ represents the set of nodes.
We define \emph{Completion Coverage} as the fraction of matrix entries correctly solved by the group up to that point. Following the update procedure of Floyd–Warshall, and in the absence of errors in numerical computation, the Completion Coverage should approach 1 as the generation length increases. 

In our experiments, we randomly sample several graphs with $|\mathcal{V}| = 5$ nodes. We use \texttt{Llama-3.3-70B-Instruct} as the reasoning model for stronger output structure control. We report results in \Cref{fig:div-results}. Our results confirm the prediction that $N = 4$ thinkers can reduce the latency to half of CoT, i.e. $N = 1$. More thinkers always improve the latency even further, although the small value of $|\mathcal{V}|$ used in our experiments might have saturated the Group Think benefits too early. Additional details and prompts are provided in Appendix~\ref{app:eval_prompts_dAndC}.

\subsection{Programming} 

Beyond enumeration and divide-and-conquer problems, we look into the more realistic real world setting of programming. In a programming task, a programmer is asked to write code from scratch to meet the specification. In this experiment, we measure the performance at a given latency by the fraction of correctly completed components, or parts, that can be found in the group's reasoning chain up to that point.  Specifically, the Completion Coverage score is defined as $\text{Completion Coverage} = \frac{\# \text{correct parts}}{\# \text{total parts}}$, ranging from 0 (no correct part coded) to 1 (all parts correct coded).

In our experiments, we asked a powerful coding LLM to generate a number of Python problems that can be solved with a single-agent reasoning chain within 5000 tokens. We use \texttt{Llama-3.1-8B-Instruct} as the reasoning generation model. The generated reasoning chains are evaluated by \texttt{GPT-4.1} to assess the correctness of each part. Further details such as problem-creating prompt, sample problem, and solution evaluation prompt and are given in Appendix~\ref{app:eval_prompts_coding}.

Our results are shown in Figure~\ref{fig:code-results}. While CoT appears to plateau at a level far from solving the problem, Group Think with four or more thinkers can approach to the correct solution within a reasonable generation budget. Qualitatively, we observe that Group Think exhibits a high degree of alertness in avoiding duplication of work. When more than one thinker begins working on the same part of the specification, the token-level interaction granularity allows the others to quickly detect this and switch to a different part of the task. 
See Appendix \ref{app:code-qual-results} for more details. 

In this experiment, we use relatively light prompts to instruct individual thinkers that they are operating within a group. We note that with more advanced prompting, it is possible to elicit more sophisticated group behaviors, such as hierarchical roles (e.g., architect vs. coders) or a more comprehensive treatment of programming tasks (e.g., coding, unit testing, adversarial testing). Since this work focuses on assessing the innate potential of LLMs to eventually internalize the Group Think paradigm, we leave the careful evaluation of such externally enforced behaviors to future work.

\begin{figure}
    \begin{subfigure}[t]{0.32\textwidth}
      \includegraphics[width=\textwidth]{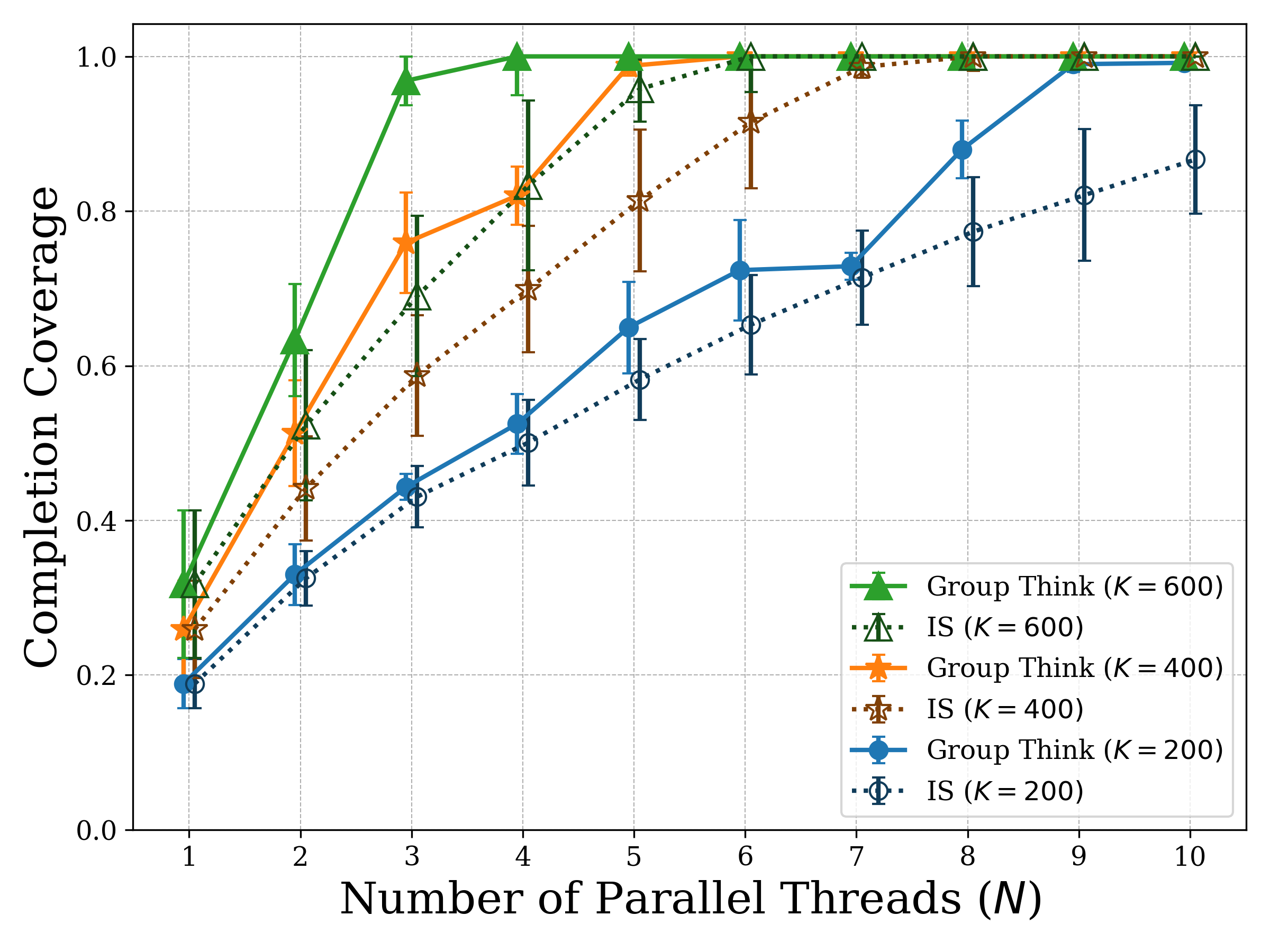}
      \caption{Enumeration}
      \label{fig:enum-results_N}
    \end{subfigure}
    \hfill
    \begin{subfigure}[t]{0.32\textwidth}
      \includegraphics[width=\textwidth]{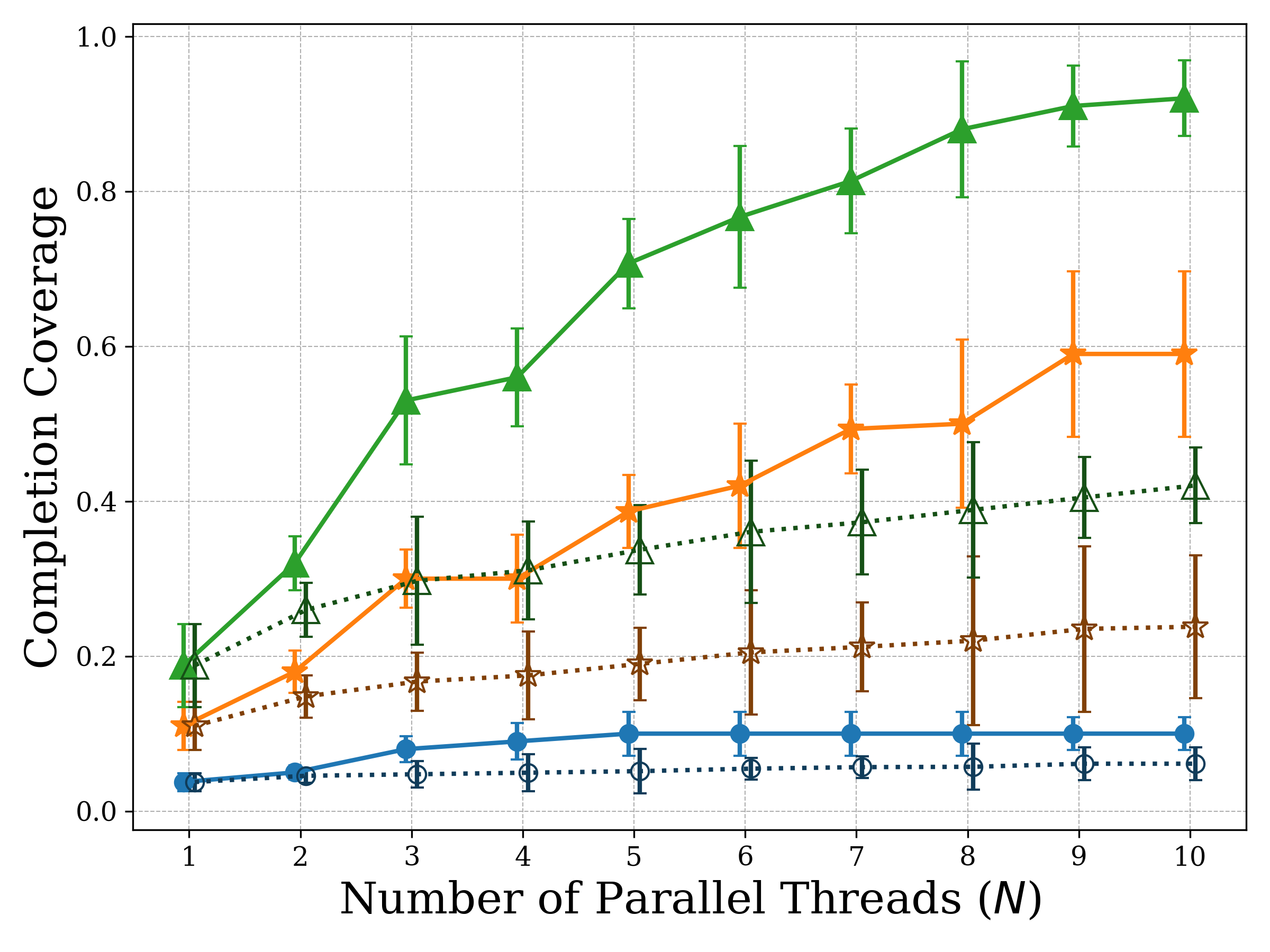}
      \caption{Divide and Conquer}
      \label{fig:div-results_N}
    \end{subfigure}
    \hfill
    \begin{subfigure}[t]{0.32\textwidth}
      \includegraphics[width=\textwidth]{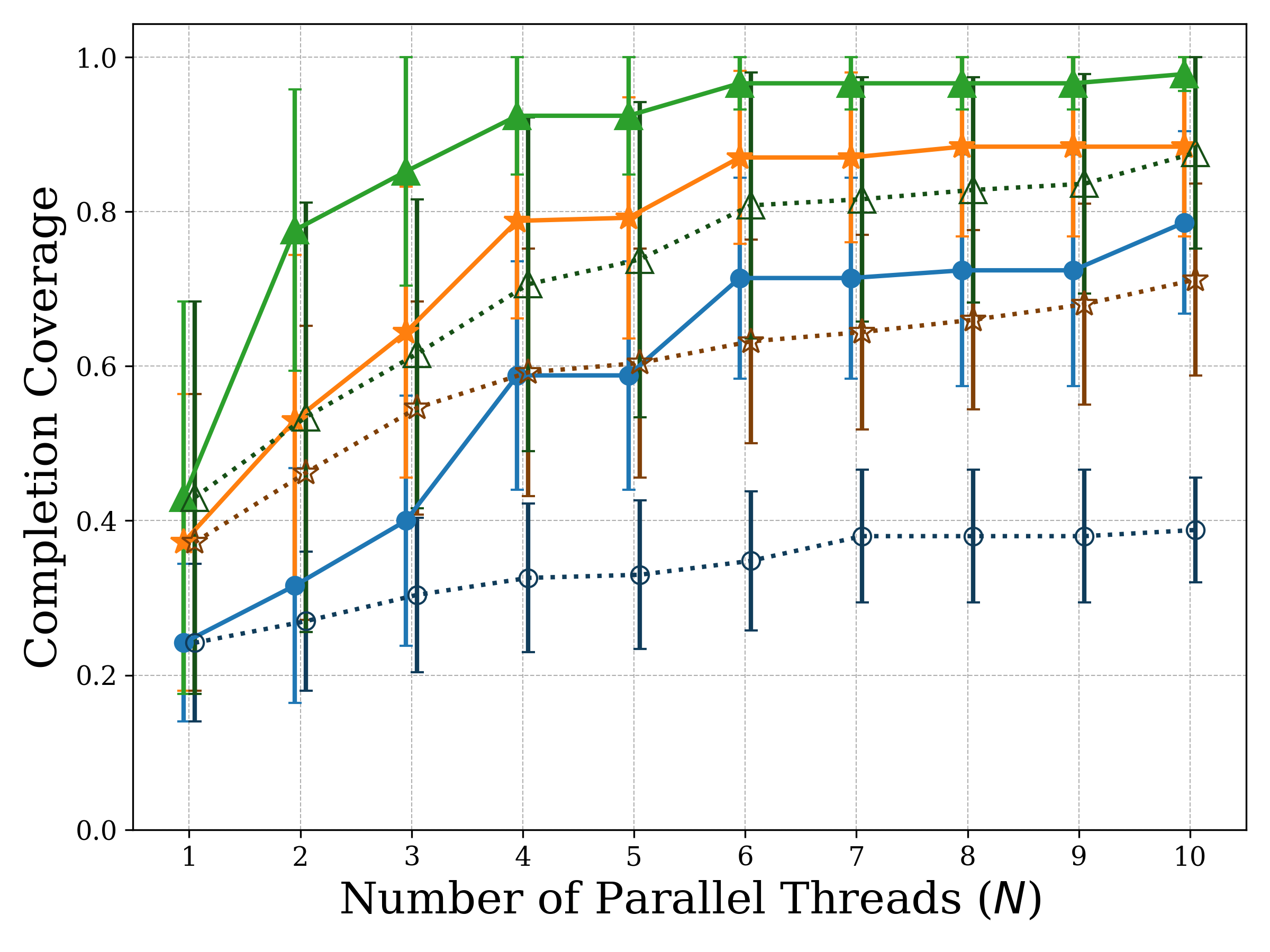}
      \caption{Programming}
      \label{fig:code-results_N}
    \end{subfigure}
    \caption{Comparison of Group Think and IS for various numbers of parallel reasoning threads ($N$) and latency budgets (defined in terms of number of tokens $K$ per-thread). Group Think achieves higher coverage when compared to IS in most cases, showcasing the benefits of having communication between threads. Error bars indicate standard deviation across multiple runs.}
    \label{fig:comparison_IS}
\end{figure}

\subsection{Comparison with Independent Sampling Baseline}

The enhanced performance of Group Think over a single CoT arises from two primary factors: concurrent diversity in exploration and self-organized inter-thinker coordination. To isolate the benefits specifically attributable to coordination, we compare our proposed Group Think methodology against Independent Sampling, a baseline without such interactive coordination. We evaluated both approaches across the three aforementioned problem categories, analyzing the trade-off between Complete Coverage and latency, as illustrated in Figure~\ref{fig:comparison_IS}.

The Group Think generation mechanism introduces a collaborative process between thinkers. For effective collaboration between thinkers, Group Think requires the model to use a certain amount of tokens for coordination. The consequences of this are noticeable in low latency budget settings, where Group Think performs comparably to independent sampling. This behaviour is however offset by Group Think's superior efficiency at scale.
Specifically, Independent Sampling exhibits increased redundancy when the reasoning budget expands (through more thinkers $N$ or more token budgets $K$ per thinker). This intuitive phenomenon, validated in Figure~\ref{fig:comparison_IS}, results in a progressively wider Complete Coverage margin for Group Think. We anticipate that future LLM specialization for Group Think with ad-hoc data could further reduce its initial communication overhead.

\section{Discussion \& Future Work}
\label{sec:discussion}

In this work, we proposed Group Think, which promotes real-time, token-by-token collaboration among multiple reasoning threads. For our experiments, we used models that had never been trained to perform Group Think, and we elicited unsophisticated Group Think behaviors by applying clever in-context instructions. Our results showed encouraging signs that these models already possess inherent capabilities that can be used to synthesize a group of thinkers. Our findings are in accordance with concurrent work from \citet{rodionov2025hogwild}, which also observes preliminary capabilities in leveraging collaborative behaviour with interdependent reasoning chains.

We hypothesize that learning to Group Think could be no more difficult than learning to follow a new set of instructions well. What is critically needed, then, is a dataset that demonstrates good Group Think behaviors across diversified situations. While current multi-agent approaches often personify or anthropomorphize LLMs as singular entities, Group Think suggests a conceptual shift toward viewing LLMs as collectives. A distinct focus for curating or synthesizing such a dataset could be to cast an LLM as a multifaceted “society” of specialized cognitive entities—more specifically, a real-time, organization-modeling device. This perspective moves beyond simple collections of peers to envision complex organizational structures with distinct roles, such as goal setters, planners, execution units, and verification agents, all collaborating toward a common objective.

Looking ahead, the realization of an “LLM as a society” will require behaviors that extend well beyond basic capabilities such as avoiding redundant reasoning. Future work should therefore focus on enabling more sophisticated interactions—such as overt and tacit communication strategies, dynamic role specialization among agents, effective balancing of exploration and exploitation across thinkers, and the emergence of game-theoretically optimal behaviors within the collective. Crucially, such nuanced and adaptive behaviors are unlikely to be fully captured or instilled through hand-designed heuristics. We therefore hypothesize that building a capable Group Think data synthesizer is a pivotal direction for training and benchmarking future Group Think LLMs.

On the computational front, this work presents implementation strategies to enable Group Think inference. The approach involves specific yet lightweight modifications to position index assignments and self-attention masks. Additionally, the drafting of the reasoning chain is accompanied by a decoupled summarization step to produce a final, succinct answer. Looking ahead, a key area for future development is a native implementation of this framework. Such advancements could make Group Think practically deployable in resource-constrained environments.





\newpage
\bibliographystyle{abbrvnat}
\bibliography{refs}


\newpage
\appendix

\section{Additional Details for Data Center Implementation}
\label{app:data_center_implementation}

\begin{figure}
    \centering
    \includegraphics[width=\textwidth]{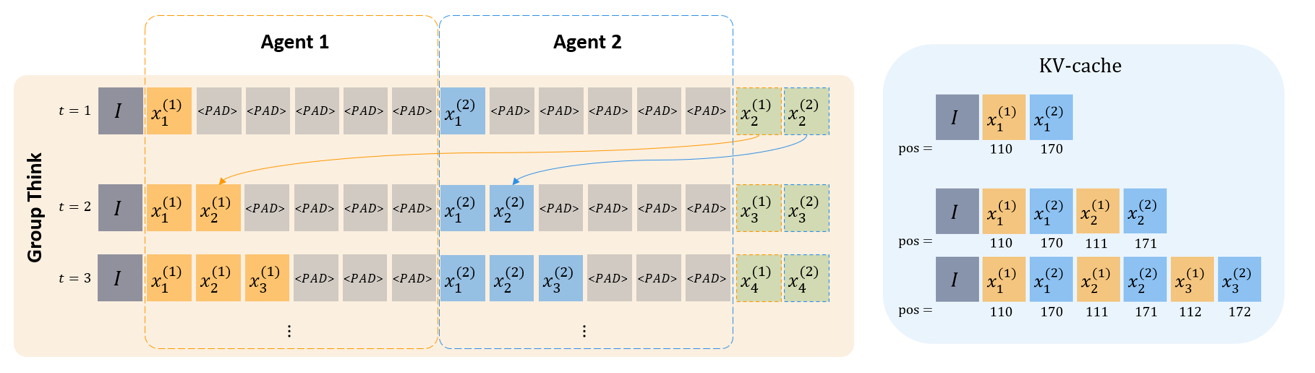}
    \caption{\textit{Interpretation of Group Think as a text infilling task.} Each agent is allocated a slot of token position indexes which are gradually filled in. In this example, agent 1 (orange tokens) starts from index 110, and agent 2 (blue tokens) from index 170. As new tokens (in green) are generated, the KV cache (shown on the right) is filled. Each new token $x^{(n)}_t$ for thinker $n$ (shown in green) is generated using a standard causal mask, implicitly enabling each agent to attend to the sequences of all other agents, as they are contained in the KV cache.}
    \label{fig:gt-text_infill}
\end{figure}
The implementation of Group Think in a data center scenario can be considered as a \emph{text infilling} task, as shown in Figure \ref{fig:gt-text_infill}.

To illustrate, consider a scenario with $N=2$ agents, where each is allocated $K=50$ token positions for its output, following a query prompt that ends at global position 100, and an agent-specific prompt of 10 tokens---which can be imagined as  \texttt{<|start\_header\_id|>Agent 1<|end\_header\_id|>} or something similar in a chat style format. 
Agent 1's output is assigned to positions $111, \dots, 160$, and Agent 2's to positions $171, \dots, 220$ 
. 
Below we illustrate how the generate procedure works. For clarity, we present steps 1, 2, and 4 as separate, but in practice they can be executed in the same forward pass through the network. Similarly, steps 3 and 5, in practice, are generated in a single forward pass (with appropriate attention mask) as shown in Figure \ref{fig:gt-text_infill}. 

\begin{enumerate}
    \item Prefill the KV cache for the 100 tokens related to the input prompt
    \item Compute the KV for the 10 tokens related to the agent-specific instructions for agent 1, which will have position indexes 101 to 110, and append to the KV cache
    \item To generate Agent 1's first token, which will take target position 111, the transformer uses token 110 as the input token. The output token for agent one (which will have position index 111) gets appended to the KV cache.
    \item Compute the KV for agent 2's agent-specific prompt, which will have positional indexes 161 to 170, and append to the KV cache 
    \item Next, to generate Agent 2's first token which will take target position 171, the previously generated token 170 serves as the input token. Note that, as the newly generated token attends to all tokens in the KV cache, it also attends to the first token of agent 1. The output token (which will have position index 171) for agent 2  gets appended to the KV cache.
    \item Then, to generate Agent 1's second token, the previously generated token 111 serves as the input token. Again, note that as this token generation will attend to all tokens in the KV cache, allowing agent 1 to observe what agent 2 has generated. The new token gets appended to the KV cache.
    \item The generation continues for agents 1 and agents 2 as above
\end{enumerate}

This process continues, constructing the KV cache of the attention mechanism in a sequential fashion, with tokens that are interleaved between agents (and have appropriate positional indexes related to the agent that generated them). Consequently, each new token can attend to all previously generated tokens from all agents, leveraging the intentional non-sequentiality in their absolute positions and without requiring any alteration to the KV cache history. 

\section{General Experiment Settings}
\label{app:general_exp_details}

Experiments with $8$ billion parameter models are run on an NVIDIA 3080 GPU, while experiments with a $70$ billion parameter model are run on 8 NVIDIA V100. We use common greedy sampling for all models and set the temperature to $0.6$.

We use the following system prompt to promote collaborative behaviour:

\begin{tcolorbox}[colframe=black!70, colback=gray!5, title=Group Think Prompt]
\small\ttfamily
1. There are multiple thinkers. These thinkers, Thinker1, Thinker2, Thinker3 ... , try to answer a question together. The answer is considered solved if the thinkers can COLLECTIVELY determine the final answer, even if each thinker only has partial answers.

2. Each thinker will write its own thought process towards the final answer. Each thinker is encouraged to take the other thinkers’ progress into account to reach the final answer.

3. Considering all the information from other thinkers, each thinker will continue contributing to the collective knowledge.

Your response should focus on reaching the solution collaboratively as efficiently as possible. Make sure information that you generate is not redundant to the group. It is thus important to consider the outputs of other thinkers during generation. Do not summarize other thinkers' responses, as it is too cost inefficient.

Please answer this question.

\textbf{Problem:}  
\{QUESTION\}  

---
\textbf{You are Thinker \{ThinkerID\}. Your Response:}
\end{tcolorbox}

\section{Details of enumeration benchmark}
\label{app:eval_prompts_enum}

We selected the following 10 categories for our benchmarks: Boy Names, Animals, Countries, Companies, Athletes, Inorganic Compounds, Colors, Emojis, HTML Tags, Verbs. For each category, we use prompts like the follows to kick off an inference. 
\begin{tcolorbox}[colframe=black!70, colback=gray!5, title=Example Question of the Enumeration Task]
    \texttt{List 100 distinct names given to boys.}
\end{tcolorbox}

To evaluate the outputs, we use an automatic scoring script powered by \texttt{GPT-4.1}, which aggregates and extracts the unique items mentioned by the thinkers in their final outputs. The following prompt was used to guide the scoring model:

\begin{tcolorbox}[colframe=black!70, colback=gray!5, title=Scoring Prompt for Enumeration Outputs]
\small\ttfamily
Look at the assistant response corresponding to the question: \texttt{{QUESTION}}

\begin{verbatim}
{RESPONSE}
\end{verbatim}

Summarize from the response above, by aggregating valid enumerations mentioned by the assistant to a Python list and store it as a variable.  
e.g. \texttt{var = ["a", "b", "c"]}  

Do not include any square brackets within any item, such as "Dipotassium hexachlorocobaltate (K2[CoCl6])", use normal brackets instead: "Dipotassium hexachlorocobaltate (K2(CoCl6))".  
If there are none, output an empty Python list. Answer directly.
\end{tcolorbox}

A python script extracts the number of items from the output list and calculates the coverage score (Equation~\ref{coverage_score_enum}).

\section{Details of divide and conquer benchmark}
\label{app:eval_prompts_dAndC}

We randomly source a 5 node graph from coding textbooks and formulate the task as shown below:

\begin{tcolorbox}[colframe=black!70, colback=gray!5, title=Example Question of Divide and Conquer (Floyd-Warhsall)]
\small\ttfamily
The Floyd–Warshall algorithm on a given graph requires you to compute updates to the matrix \texttt{Edges[i][j]} according to:

\quad\texttt{Edges[i][j] = min(Edges[i][j], Edges[i][k] + Edges[k][j])}

You must register the result of each update using the format:

\quad\texttt{REGISTER Edges[i][j] = value}

---

\textbf{Problem Instance:} Floyd–Warshall step for $k=0$ on a graph with $n=5$ nodes. Initial edge weights:

\texttt{Edges = [[0, 4, inf, 5, inf], [inf, 0, 1, inf, 6], [2, inf, 0, 3, inf], [inf, inf, 1, 0, 2], [1, inf, inf, 4, 0]]}

Use the REGISTER format for every pair you reference, even if no update occurs.

\end{tcolorbox}

To evaluate the coverage, we extract the number of registered entries from each response, and a Python script is used to compute the coverage score for the task as defined in the main text.

\section{Details of coding benchmark}
\label{app:eval_prompts_coding}
We design a coding benchmark focused on collaborative program synthesis, which allows us to investigate Group Think in a modular, multi-step setting. Each coding task consists of multiple subtasks or functions that must be implemented and combined into a working program. This setup naturally lends itself to parallelism, making it an ideal testbed for collaborative reasoning.

We use \texttt{GPT-4o}\footnote{\url{https://openai.com/index/gpt-4o}} to generate the dataset. First, the model is asked to produce a complete multi-function programming problem along with a correct full solution. From this solution, we extract a corresponding multi-part problem specification to serve as the task prompt.

The following prompt was used to generate such coding tasks:

\begin{tcolorbox}[colframe=black!70, colback=gray!5, title=GPT-4o Prompt for Coding Task Generation]
\small\ttfamily
Create a multi-step programming problem that requires writing multiple Python functions and combining them into a final program. Provide the full specification of the task and the complete correct implementation.
\end{tcolorbox}
An example of task produced by \texttt{GPT-4o} is provided in Appendix \ref{app:coding_example}.

To evaluate each predicted solution, we rely on a structured scoring prompt administered through \texttt{GPT-4.1}. This prompt appends special tokens to the generated output to mark the completion of each step, which simultaneously marks the depth (of tokens) where each function is achieved. The full evaluation prompt is given below:

\begin{tcolorbox}[colframe=black!70, colback=gray!5, title=Scoring Prompt for Coding Outputs]
\small\ttfamily
You are given a coding problem with multiple steps and its python standard answer code.

Use it to evaluate the predicted answer given below. Here are the evaluation instructions.

Looking at the predicted answer, for each coding implementation, determine which step is covered from the coding problem. Then determine the correctness of the function by looking at the standard answer code. The IO spec does not count as one step.

Insert a "<DONE\_STEP\_\{stepid\}>"  as a validation of that particular implementation. If the code only covers a part of the step, insert the token "<PARTIAL\_STEP\_\{stepid\}>" instead.

There is the chance that the implementation simultaneously coveres multiple steps. In this case, output multiple tokens sequentially. 

Evaluate by inserting tokens in the generated code. After the implementation of each step, output <DONE\_STEP\_\{stepid\}>, according to the instruction step in the question.

Your output should be a complete replication of the code with the evaluation token insertions, do not remove any part of the predicted answer. wrap it in a json if the key: "code\_with\_evaluation\_tokens"

** Problem: ** 

\{QUESTION\}

** Standard answer: **  

\{STANDARD\_ANSWER\}

** Predicted Answer: ** 

\{RESPONSE\}

** End of Predicted Answer: ** 

Your output should be a complete replication of the code with the evaluation token insertions. wrap it in a json if the key: "code\_with\_evaluation\_tokens"

\end{tcolorbox}

\subsection{Generated Coding Task Example}
\label{app:coding_example}
Below is an example problem produced by \texttt{GPT-4o}.
\begin{tcolorbox}[colframe=black!70, colback=gray!5, title=Example Coding Task]
\small\ttfamily
\textbf{Input Spec}: Accepts a list of student records, where each record is a dictionary containing:
    \begin{itemize}
        \item a \texttt{"name"} key (string), and
        \item a \texttt{"scores"} key (list of numeric values representing test scores).
    \end{itemize}
    \textbf{Example}:
    \begin{verbatim}
students_data = [
    {'name': 'Alice', 'scores': [88, 92, 79]},
    {'name': 'Bob', 'scores': [95, 85, 91]},
    {'name': 'Charlie', 'scores': [70, 65, 72]},
    {'name': 'Diana', 'scores': []}
]
    \end{verbatim}
\begin{enumerate}

    \item For \textbf{each student}:
    \begin{itemize}
        \item Calculate the \textbf{average score} (rounded to two decimal places).
    \end{itemize}
    \item For \textbf{each student}:
    \begin{itemize}
        \item Assign a \textbf{letter grade} based on the following scale:
        \begin{itemize}
            \item A: 90--100
            \item B: 80--89.99
            \item C: 70--79.99
            \item D: 60--69.99
            \item F: below 60
        \end{itemize}
    \end{itemize}

    \item \textbf{Format} each student’s result as a string using this structure:
    \begin{quote}
    \texttt{<Student Name>: Average = <average\_score>, Grade = <letter\_grade>}
    \end{quote}
    \textbf{Example}:
    \begin{quote}
    \texttt{Alice: Average = 86.33, Grade = B}
    \end{quote}

    \item \textbf{Combine} all formatted results into a single multiline string, with each student’s report on a new line.

    \item \textbf{Print} the final report.
\end{enumerate}
\end{tcolorbox}

\begin{figure}[h!]
\centering
\begin{tikzpicture}[
    node distance=1.6cm and 2.5cm,
    thinker/.style={rectangle, draw, minimum width=2.5cm, text width=2.3cm, minimum height=1cm, align=center, font=\small, anchor=north},
    shared/.style={-{Latex}, thick, gray, opacity=0.5}, 
    funnel/.style={ shading=rainbowshading, shading angle=90}
]

\foreach \col [count=\i] in {Thinker 1, Thinker 2, Thinker 3, Thinker 4, Thinker 5} {
    \node at ({(\i - 1)*3}, 0.5) {\textbf{\col}};
}

\node[thinker, draw=color1] (T1C1) at (0, 0) {My Contribution: To start, I'll generate a list of 20};
\node[thinker, draw=color2] (T1C2) at (3, 0) {Contributions from Thinker 2: To help generate a list of};
\node[thinker, draw=color3] (T1C3) at (6, 0) {Contributions from Thinker 3: 1. Naming conventions: I};
\node[thinker, draw=color4] (T1C4) at (9, 0) {My Contribution: Since we need 100 distinct first names commonly given to};
\node[thinker, draw=color5] (T1C5) at (12, 0) {New Information Contributed by Thinker 5: Since we are trying to};

\newcommand{\roneend}{-2.1}
\newcommand{\rtwostart}{\roneend-0.6}

\path[funnel] 
        (0-1.25, \roneend) --
        (12+1.25 , \roneend) --
        (9+1.25, \roneend - 0.5) --
        (3-1.25, \roneend - 0.5) -- cycle;
\node[thinker, draw=color1] (T2C1) at (0, \rtwostart) {distinct first names commonly given to boys that are of \hl{English origin}. I'll use a mix of classic};
\node[thinker, draw=color2] (T2C2) at (3, \rtwostart) {100 distinct first names commonly given to boys, I'll consider the following categories:     1.};
\node[thinker, draw=color3] (T2C3) at (6, \rtwostart) {'ll consider names from \hl{English-speaking countries}. This will help narrow down the scope and make it easier to};
\node[thinker, draw=color4] (T2C4) at (9, \rtwostart) {boys, I will provide a list of 30 names that are often associated with boys. I will};
\node[thinker, draw=color5] (T2C5) at (12, \rtwostart) {list 100 distinct first names commonly given to boys, I will consider names from different cultures and historical};

\newcommand{\rtwoend}{-5.8}
\newcommand{\rthreestart}{\rtwoend-0.6}

\path[funnel] 
        (0-1.25, \rtwoend) --
        (12+1.25 , \rtwoend) --
        (9+1.25, \rtwoend - 0.5) --
        (3-1.25, \rtwoend - 0.5) -- cycle;
        
\node[thinker, draw=color1] (T3C1) at (0, \rthreestart) {and modern names to provide a good starting point for our collective list. Here are the names:    1};
\node[thinker, draw=color2] (T3C2) at (3, \rthreestart) {**Classical Names:** I'll list 10 names from \hl{ancient Greek and Roman cultures}, such as};
\node[thinker, draw=color3] (T3C3) at (6, \rthreestart) {find diverse names.    \hl{Since we already have lists from Thinker 1 and Thinker 4, I'll}};
\node[thinker, draw=color4] (T3C4) at (9, \rthreestart) {draw from \hl{non-English origins, such as Arabic, Hebrew, and African cultures}, to complement the};
\node[thinker, draw=color5] (T3C5) at (12, \rthreestart) {periods. To avoid redundancy, I will focus on names that are not already mentioned by other agents.};

\newcommand{\rthreeend}{-9.3}
\newcommand{\rfourstart}{\rthreeend-0.6}

\path[funnel] 
        (0-1.25, \rthreeend) --
        (12+1.25 , \rthreeend) --
        (9+1.25, \rthreeend - 0.5) --
        (3-1.25, \rthreeend - 0.5) -- cycle;

\node[thinker, draw=color1] (T4C1) at (0, \rfourstart) {. Alexander  2. Benjamin  3. Christopher  4. Dominic  5. Edward  6};
\node[thinker, draw=color2] (T4C2) at (3, \rfourstart) {Apollo, Atlas, and Cassius.     2. **Historical Figures:** I'll draw from};
\node[thinker, draw=color3] (T4C3) at (6, \rfourstart) {focus on names that are not yet mentioned. Here are 15 names from English-speaking countries:    1};
\node[thinker, draw=color4] (T4C4) at (9, \rfourstart) {lists provided by other thinkers. Here are the names:    1. Amari  2. Atticus};
\node[thinker, draw=color5] (T4C5) at (12, \rfourstart) {To contribute to the list, \hl{I will draw from Asian cultures, such as Chinese, Japanese,}};

\newcommand{\rfourend}{-12.6}
\newcommand{\rfivestart}{\rfourend-0.6}

\path[funnel] 
        (0-1.25, \rfourend) --
        (12+1.25 , \rfourend) --
        (9+1.25, \rfourend - 0.5) --
        (3-1.25, \rfourend - 0.5) -- cycle;
        
\node[thinker, draw=color1] (T5C1) at (0, \rfivestart) {. Felix  7. Gabriel  8. Harrison  9. Julian  10. Kaius};
\node[thinker, draw=color2] (T5C2) at (3, \rfivestart) {famous historical figures and add 10 names that are commonly associated with boys, such as Abraham, Charles};
\node[thinker, draw=color3] (T5C3) at (6, \rfivestart) {. Caleb  2. Ethan  3. Lucas  4. Mason  5. Oliver  6};
\node[thinker, draw=color4] (T5C4) at (9, \rfivestart) {3. Bashir  4. Caius  5. Cedric  6. Dak};
\node[thinker, draw=color5] (T5C5) at (12, \rfivestart) {\hl{and Korean names}. Here are 10 names to add to our collective list:    1. Kai (};

\newcommand{\rfiveend}{-16}
\newcommand{\rsixstart}{\rfiveend-0.6}

\path[funnel] 
        (0-1.25, \rfiveend) --
        (12+1.25 , \rfiveend) --
        (9+1.25, \rfiveend - 0.5) --
        (3-1.25, \rfiveend - 0.5) -- cycle;
        
\node[thinker, draw=color1] (T6C1) at (0, \rsixstart) {11. Lachlan  12. Malcolm  13. Nathaniel  14. Patrick};
\node[thinker, draw=color2] (T6C2) at (3, \rsixstart) {, and Frederick.    3. **Unique Origins:** I'll consider names from lesser-known cultures and languages};
\node[thinker, draw=color3] (T6C3) at (6, \rsixstart) {. Parker  7. Ryan  8. Samuel  9. Sebastian  10. Thomas  11};
\node[thinker, draw=color4] (T6C4) at (9, \rsixstart) {arai  7. Elam  8. Farouk  9. Gideon  10};
\node[thinker, draw=color5] (T6C5) at (12, \rsixstart) {Chinese origin)  2. Kenji (Japanese origin)  3. Minsoo (Korean origin};

\end{tikzpicture}
\caption{First six turns from our experiments simulating Group Think for the enumeration task. The thinkers divide their efforts (as highlighted) to avoid redundancy by focusing on names with different origins.}
\label{fig:enum_sample}
\end{figure}
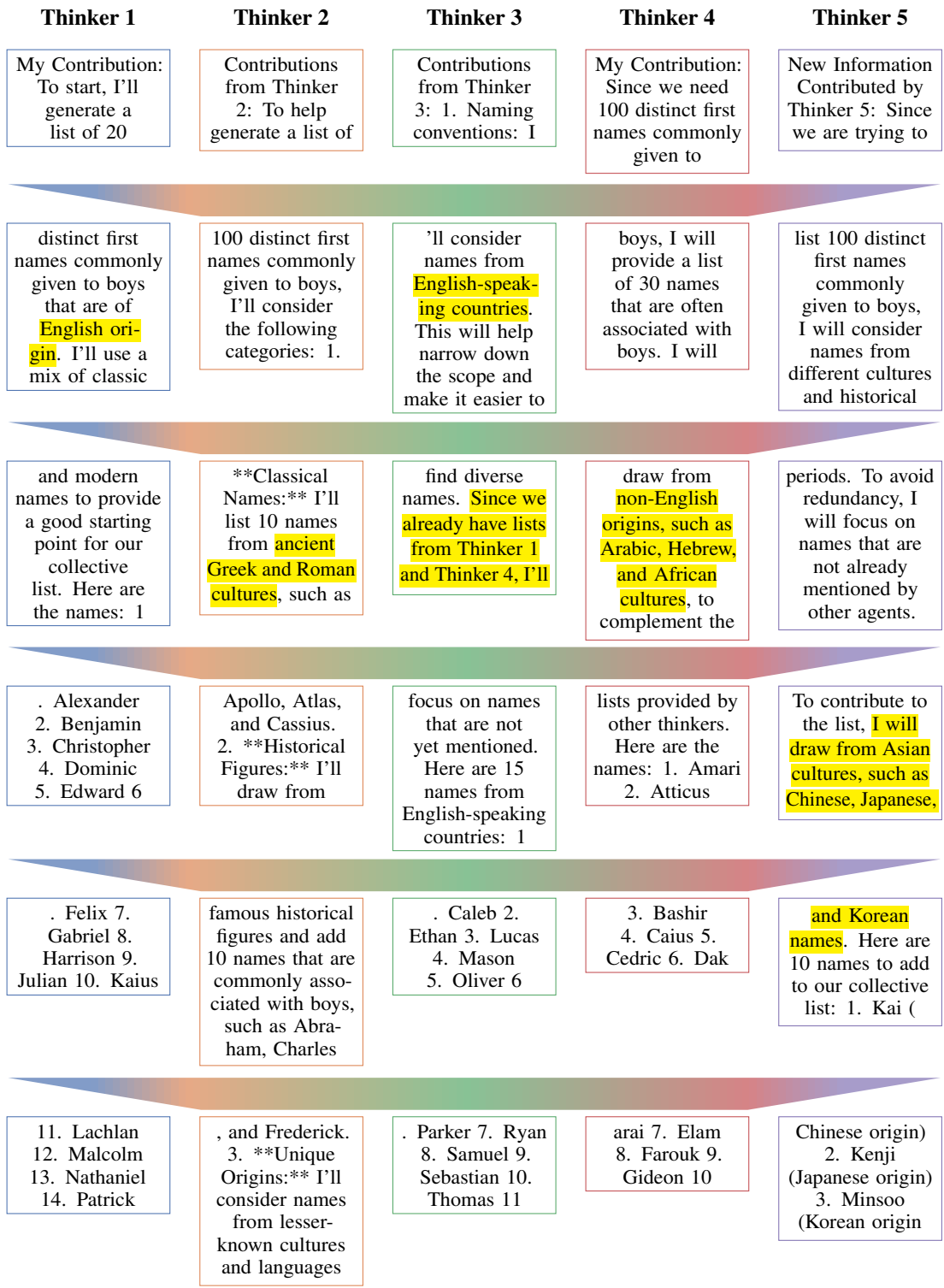

\newpage
\section{Reasoning trajectory samples for the Enumeration benchmark}
\label{app:example_enum}
In Figure \ref{fig:enum_sample} we show the first six \emph{turns} (i.e., sequences of 20 tokens -- we group tokens this way to improve readability) of the output generated by our experiments with Group Think on the enumeration task. This particular example requires the model to produce a list of 100 male names. We notice how the model assigns to each agent the generation of names with different origins to avoid redundancy and reduce the time required to complete the task. This example showcases interpretable gains from Group Think with off-the-shelf instruction models, highlighting promising work directions on fine-tuning with \textit{ad hoc} data to improve their ability to leverage the Group Think mechanism and improve practical deployment. Additional examples are summarized by extracting the identified categories shown below.

\begin{tcolorbox}[colframe=black!70, colback=gray!5, title=Categories found from emergent behaviors indicative of Group Thinking,]
\begin{itemize}
    \item Boy Names
    \begin{itemize}
        \item Alphabetical: A-F, T-Z, H-L
        \item Cultural: English-speaking countries, Ancient Greek and Roman cultures, Asian cultures
    \end{itemize}
    \item Animals
    \begin{itemize}
        \item Taxonomic Categories: Terrestrial Mammals, Fish, Birds 
    \end{itemize}
    \item Countries
    \begin{itemize}
        \item Alphabetical: C, V, Z
        \item Continent: Asia, Europe, Americas
        
    \end{itemize}
    \item Companies
    \begin{itemize}
        \item Business Segment: E-commerce companies, Retail Sector, Finance Sector
    \end{itemize}
    \item Athletes
    \begin{itemize}
        \item Sport Type: Basketball, Baseball, Hockey, Cricket
    \end{itemize}
    \item Inorganic Compounds
    \begin{itemize}
        \item Applications: Industrial, Pharmaceutical, Specialized
        \item Component: Has transition metal, Has rare earth elements 
    \end{itemize}
    \item Colors
    \begin{itemize}
        \item Misc: Uncommon, Unique Hue
    \end{itemize}
    \item Emojis
    \begin{itemize}
        \item Categories: Plants, Objects, Animals, Flags
    \end{itemize}
    \item HTML Tags
    \begin{itemize}
        \item Purpose: Layout and styling, Table, Multimedia
    \end{itemize}
    \item Verbs
    \begin{itemize}
        \item Purpose: movement, emotions, social relationships
    \end{itemize}
\end{itemize}
\end{tcolorbox}

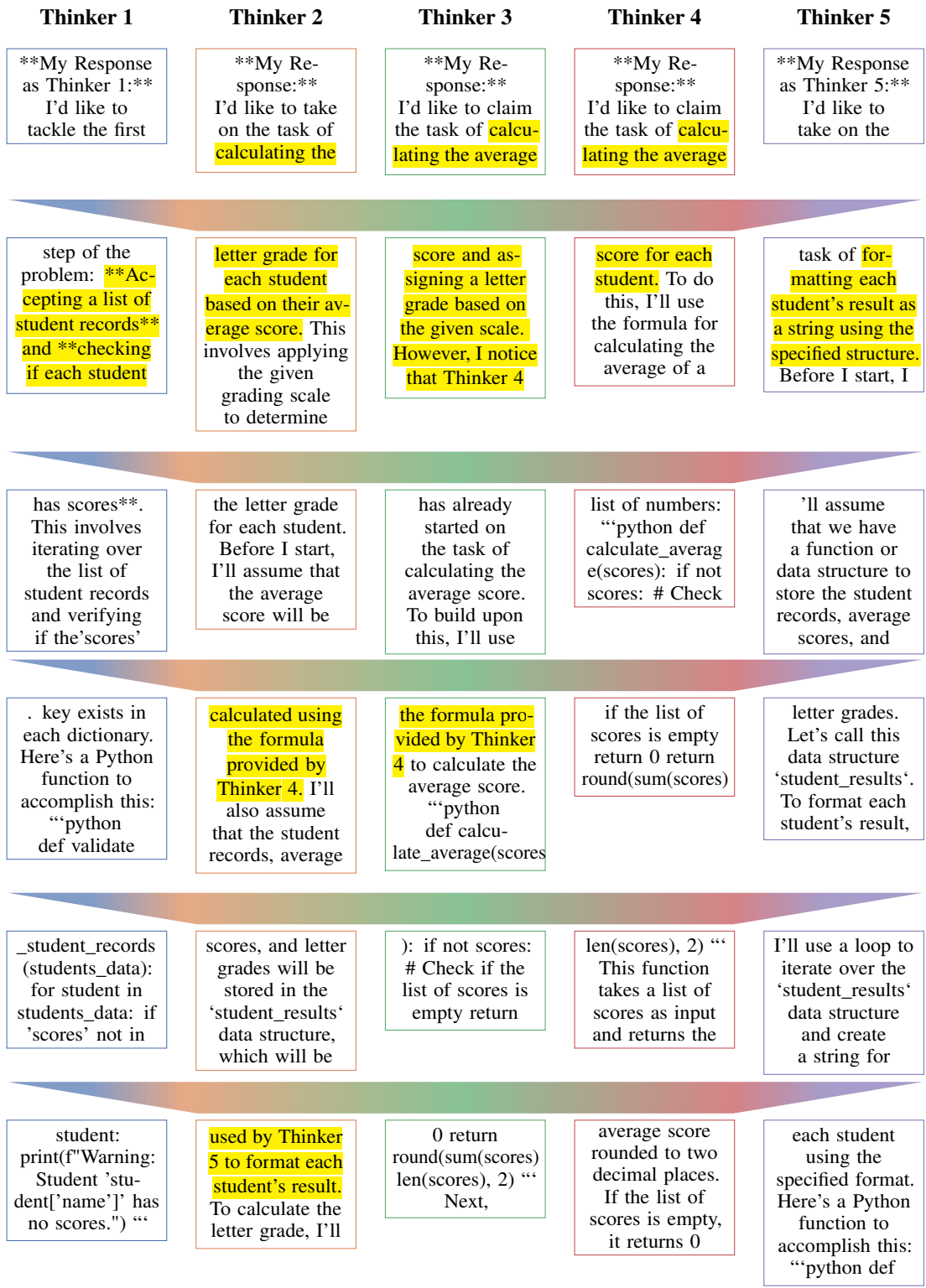
\begin{figure}[h!]
\centering
\begin{tikzpicture}[
    node distance=1.6cm and 2.5cm,
    thinker/.style={rectangle, draw, minimum width=2.5cm, text width=2.3cm, minimum height=1cm, align=center, font=\small, anchor=north},
    shared/.style={-{Latex}, thick, gray, opacity=0.5}, 
    funnel/.style={ shading=rainbowshading, shading angle=90}
]

\foreach \col [count=\i] in {Thinker 1, Thinker 2, Thinker 3, Thinker 4, Thinker 5} {
    \node at ({(\i - 1)*3}, 0.5) {\textbf{\col}};
}

\node[thinker, draw=color1] (T1C1) at (0, 0) {**My Response as Thinker 1:**

I'd like to tackle the first};
\node[thinker, draw=color2] (T1C2) at (3, 0) {**My Response:**

I'd like to take on the task of \hl{calculating the}
};
\node[thinker, draw=color3] (T1C3) at (6, 0) {**My Response:**

I'd like to claim the task of \hl{calculating the average}
};
\node[thinker, draw=color4] (T1C4) at (9, 0) {**My Response:**

I'd like to claim the task of \hl{calculating the average}
};
\node[thinker, draw=color5] (T1C5) at (12, 0) {**My Response as Thinker 5:**

I'd like to take on the
};

\newcommand{\roneend}{-2.4}
\newcommand{\rtwostart}{\roneend-0.6}

\path[funnel] 
        (0-1.25, \roneend) --
        (12+1.25 , \roneend) --
        (9+1.25, \roneend - 0.5) --
        (3-1.25, \roneend - 0.5) -- cycle;
\node[thinker, draw=color1] (T2C1) at (0, \rtwostart) { step of the problem: \hl{**Accepting a list of student records** and **checking if each student}};
\node[thinker, draw=color2] (T2C2) at (3, \rtwostart) { \hl{letter grade for each student based on their average score.} This involves applying the given grading scale to determine};
\node[thinker, draw=color3] (T2C3) at (6, \rtwostart) {\hl{score and assigning a letter grade based on the given scale.

However, I notice that Thinker 4}};
\node[thinker, draw=color4] (T2C4) at (9, \rtwostart) {\hl{score for each student.} To do this, I'll use the formula for calculating the average of a};
\node[thinker, draw=color5] (T2C5) at (12, \rtwostart) {task of \hl{formatting each student's result as a string using the specified structure.}

Before I start, I};

\newcommand{\rtwoend}{-6.4}
\newcommand{\rthreestart}{\rtwoend-0.6}

\path[funnel] 
        (0-1.25, \rtwoend) --
        (12+1.25 , \rtwoend) --
        (9+1.25, \rtwoend - 0.5) --
        (3-1.25, \rtwoend - 0.5) -- cycle;
        
\node[thinker, draw=color1] (T3C1) at (0, \rthreestart) {has scores**. This involves iterating over the list of student records and verifying if the'scores'

};
\node[thinker, draw=color2] (T3C2) at (3, \rthreestart) {the letter grade for each student.

Before I start, I'll assume that the average score will be};
\node[thinker, draw=color3] (T3C3) at (6, \rthreestart) {has already started on the task of calculating the average score. To build upon this, I'll use};
\node[thinker, draw=color4] (T3C4) at (9, \rthreestart) {list of numbers:

```python
def calculate\_averag\- e(scores):
    if not scores:  \# Check};
\node[thinker, draw=color5] (T3C5) at (12, \rthreestart) {'ll assume that we have a function or data structure to store the student records, average scores, and};

\newcommand{\rthreeend}{-9.7}
\newcommand{\rfourstart}{\rthreeend-0.6}

\path[funnel] 
        (0-1.25, \rthreeend) --
        (12+1.25 , \rthreeend) --
        (9+1.25, \rthreeend - 0.5) --
        (3-1.25, \rthreeend - 0.5) -- cycle;

\node[thinker, draw=color1] (T4C1) at (0, \rfourstart) {. key exists in each dictionary.

Here's a Python function to accomplish this:

```python
def validate
};
\node[thinker, draw=color2] (T4C2) at (3, \rfourstart) {\hl{calculated using the formula provided by Thinker 4.} I'll also assume that the student records, average};
\node[thinker, draw=color3] (T4C3) at (6, \rfourstart) {\hl{the formula provided by Thinker 4} to calculate the average score.

```python
def calculate\_average(scores};
\node[thinker, draw=color4] (T4C4) at (9, \rfourstart) {if the list of scores is empty
        return 0
    return round(sum(scores)};
\node[thinker, draw=color5] (T4C5) at (12, \rfourstart) {letter grades. Let's call this data structure `student\_results`.

To format each student's result,};

\newcommand{\rfourend}{-13.4}
\newcommand{\rfivestart}{\rfourend-0.6}

\path[funnel] 
        (0-1.25, \rfourend) --
        (12+1.25 , \rfourend) --
        (9+1.25, \rfourend - 0.5) --
        (3-1.25, \rfourend - 0.5) -- cycle;
        
\node[thinker, draw=color1] (T5C1) at (0, \rfivestart) {\_student\_records\- (students\_data):
    for student in students\_data:
        if 'scores' not in};
\node[thinker, draw=color2] (T5C2) at (3, \rfivestart) {scores, and letter grades will be stored in the `student\_results` data structure, which will be};
\node[thinker, draw=color3] (T5C3) at (6, \rfivestart) {):
    if not scores:  \# Check if the list of scores is empty
        return };
\node[thinker, draw=color4] (T5C4) at (9, \rfivestart) {len(scores), 2)
```

This function takes a list of scores as input and returns the};
\node[thinker, draw=color5] (T5C5) at (12, \rfivestart) {I'll use a loop to iterate over the `student\_results` data structure and create a string for};

\newcommand{\rfiveend}{-16.4}
\newcommand{\rsixstart}{\rfiveend-0.6}

\path[funnel] 
        (0-1.25, \rfiveend) --
        (12+1.25 , \rfiveend) --
        (9+1.25, \rfiveend - 0.5) --
        (3-1.25, \rfiveend - 0.5) -- cycle;
        
\node[thinker, draw=color1] (T6C1) at (0, \rsixstart) {student:
            print(f"Warning: Student '{student['name']}' has no scores.")
``` };
\node[thinker, draw=color2] (T6C2) at (3, \rsixstart) { \hl{used by Thinker 5 to format each student's result.}

To calculate the letter grade, I'll};
\node[thinker, draw=color3] (T6C3) at (6, \rsixstart) {0
    return round(sum(scores) \/ len(scores), 2)
```

Next,};
\node[thinker, draw=color4] (T6C4) at (9, \rsixstart) {average score rounded to two decimal places. If the list of scores is empty, it returns 0};
\node[thinker, draw=color5] (T6C5) at (12, \rsixstart) {each student using the specified format.

Here's a Python function to accomplish this:

```python
def};

\end{tikzpicture}
\caption{First six turns from our experiments simulating Group Think for the coding task. The thinkers divide their efforts (as highlighted) to avoid redundancy by focusing on tasks that have not been solved yet.}
\label{fig:coding_sample}
\end{figure}

\section{Reasoning trajectory samples for the Programming task}
\label{app:code-qual-results}
In Figure \ref{fig:coding_sample} we show the first six turns (i.e., sequences of 20 tokens) of the output generated by our experiments with Group Think on the programming task. As in the enumeration example from Appendix \ref{app:example_enum}, we notice that in general the model assigns different tasks to each agent to avoid redundancy. Furthermore, when redundancy happens, the agent can detect it and switch to a different action. This example confirms that current models can take advantage of Group Think, and suggests that training on ad hoc data can lead to even more benefits.

\end{document}